\documentclass[runningheads, envcountsame, a4paper]{llncs}
\usepackage[T1]{fontenc}
\usepackage{graphicx}
\usepackage{booktabs}
\usepackage[misc]{ifsym}
\newcommand{\corr}{(\Letter)}
\usepackage{bm}
\usepackage{multirow}
\usepackage{makecell}
\usepackage{amsmath}
\usepackage[ruled,vlined]{algorithm2e}
\usepackage{subfigure}
\usepackage{xcolor}

\begin{document}

\title{AdaHAT: Adaptive Hard Attention to the Task in Task-Incremental Learning}

\titlerunning{Adaptive Hard Attention to the Task in Task-Incremental Learning}
% If the full title of your paper is short enough to also fit in the running head, you can omit the abbreviated paper title here. You can check as follows: if you comment out the \titlerunning line, something will appear in the header of all odd-numbered pages of your PDF from page 3 onward. This something is either the full title (in which case all is well), or the error message "Title Suppressed Due to Excessive Length". If this error message appears, you're going to want to provide an abbreviated title within the \titlerunning command, because if you won't do it, Springer will do it for you.

%N.B.: Author information (both in the \author{} and \authorrunning{} command) should only be present in the Camera-Ready Version of your paper. The version that you initially submit for review, ought to be double-blind. So, when initially submitting your paper, use:
%\author{Author information scrubbed for double-blind reviewing}
\author{Pengxiang Wang\inst{1}\orcidID{0009-0006-3479-2298} \and
Hongbo Bo\inst{2,3} \and
 Jun Hong\inst{4}  \and Weiru Liu\inst{3} \and Kedian Mu\inst{1} \corr        %\orcidID{0000-1111-2222-3333}
 }
% You may leave out the orcidID information, if you want to.
% Use \corr to indicate the corresponding author. Note the spacing around the \corr command. Only one author can be the corresponding author.

\authorrunning{P. Wang et al.}
% If there are two authors, write 'A.L. Benjamin and C.C. Broadus Jr.'
% If there are more than two authors, '[...] et al.' is used.

\institute{School of Mathematical Sciences, Peking University, Beijing, China \email{wangpengxiang@stu.pku.edu.cn, mukedian@math.pku.edu.cn}
\and
Population Health Sciences Institute, Newcastle University, Newcastle, UK \email{hongbo.bo@newcastle.ac.uk}
\and 
School of Engineering Mathematics and Technology, University of Bristol, Bristol, UK \email{weiru.liu@bristol.ac.uk}
\and
School of Computing and Creative Technologies, University of the West of England, Bristol, UK \email{Jun.Hong@uwe.ac.uk}
}

\tocauthor{Pengxiang~Wang,Hongbo~Bo,Jun~Hong,Weiru~Liu,Kedian~Mu}
\toctitle{AdaHAT: Adaptive Hard Attention to the Task in Task-Incremental Learning}

\maketitle              % typeset the header of the contribution

\begin{abstract}
  Catastrophic forgetting is a major problem in task-incremental learning, where neural networks tend to overwrite previously learned knowledge when trained on new tasks. A number of architecture-based approaches have been proposed to address this problem. However, the architecture-based approaches suffer from another problem related to network capacity when the networks learn long task sequences: As a network is trained on an increasing number of new tasks in a long task sequence, a growing proportion of active parameters becomes static to prevent forgetting of previously learned knowledge. In this paper, we propose Adaptive Hard Attention to the Task (AdaHAT)  with an adaptive attention mechanism which allows adaptive updates to static parameters by taking into account the information about previous tasks on both the importance of these parameters to previous tasks and the current network capacity. Based on this idea, we develop a new neural network architecture incorporating our proposed AdaHAT mechanism. AdaHAT extends an existing architecture-based approach, Hard Attention to the Task (HAT), to better support task-incremental learning over long task sequences. We conduct experiments on a number of datasets and compare AdaHAT with task-incremental learning baselines including HAT. Our experimental results show that AdaHAT achieves better average performance across tasks than these baselines, especially on long task sequences, demonstrating the benefits from balancing the trade-off between stability and plasticity of a network  when learning such sequences of tasks, alleviating the network capacity problem. Our code is available\footnote{\url{pengxiang-wang.com/projects/continual-learning-arena}}.
  

\end{abstract}

\section{Introduction}

One of the key features of human intelligence is the ability to learn continually and adapt to new information over time. One of the fundamental challenges faced by deep neural networks in continual learning is catastrophic forgetting~\cite{mccloskey1989catastrophic,ratcliff1990connectionist,goodfellow2013empirical}, which results in drastic performance degradation on previous tasks as a network is sequentially trained on new information. Continual learning, and its scenario task-incremental learning, aim to address this problem by aiming to learn and accumulate knowledge over a sequence of tasks without catastrophic forgetting~\cite{de2021continual}.

To overcome catastrophic forgetting in task-incremental learning, various strategies have been proposed. Most of these strategies adopt the idea of leveraging certain forms of information about previous tasks and incorporating them into the learning process for new tasks to prevent forgetting previously learned knowledge. For example, replay-based approaches mitigate forgetting by storing parts of previous task data, which replay algorithms use to consolidate previous knowledge~\cite{lopez2017gradient,shin2017continual,buzzega2020dark}; regularization-based approaches introduce regularization terms constructed using information about previous tasks into the loss function when training new tasks~\cite{li2017learning,kirkpatrick2017overcoming,zenke2017continual,lee2017overcoming,nguyen2017variational}.

Architecture-based approaches have been proposed to exploit the inherent nature of parameter separability within the architecture of neural networks, focusing on reducing representational overlap in the network. The core idea in these approaches is to allocate parameters in different parts of the network to different tasks, to keep the parameters learned for previous tasks from being significantly changed when learning new tasks~\cite{fernando2017pathnet,mallya2018packnet,mallya2018piggyback,serra2018overcoming,wortsman2020supermasks}. Therefore, they are also referred to as parameter isolation methods~\cite{de2021continual}. An important consideration in these approaches is the trade-off between stability and plasticity of the network, where stability is reflected  in performance over learned tasks while plasticity is reflected in saving network capacity by keeping parts of the network active for new tasks~\cite{wang2024comprehensive}.

A recent architecture-based approach to reduce representational overlap is the task-based hard attention mechanism called Hard Attention to the Task (HAT)~\cite{serra2018overcoming}. HAT learns layer-wise attention vectors (masks) for each task, concurrently to learning network parameters, to protect parameters that are important to previous tasks. When learning new tasks, HAT freezes parameters allocated to previous tasks, thus preventing forgetting what has been learned. Specifically, HAT achieves this behavior through adjusting gradients during the training process to manipulate the updates to the parameters directly. However, as the number of tasks increases, network capacity is rapidly saturated, which greatly reduces the proportion of remaining active parameters for new tasks. We refer to this as the network capacity problem. Here, HAT overemphasizes stability at the cost of reduced plasticity, while other categories of continual learning approaches typically exhibit the opposite tendency.

In this paper, we propose an extension to the task-based hard attention mechanism proposed in HAT. Following the spirit of HAT, we propose a new mechanism that introduces adaptive attention to the task, which allows adaptive updates to those static network parameters that have been allocated to previous tasks when learning new tasks, taking into account the information about previous tasks on both the importance of these parameters to previous tasks and the current network capacity. These adaptive parameter updates help to reuse, in a measured way, parts of the network that have been made static for previous tasks. We call our proposed mechanism Adaptive Hard Attention to the Task (AdaHAT). AdaHAT has three distinctive characteristics: 1. It balances the trade-off between stability and plasticity of the network, i.e., between protecting important parameters to previous tasks and preserving network capacity for new tasks; 2. It particularly suits for learning long task sequences; 3. It is an adaptive process to update previously static parameters, based on both their importance to previous tasks and network capacity usage. For instance, when the network capacity is insufficient, larger updates are allowed to parameters that have been allocated to fewer previous tasks, hence preserving more network capacity by compromising only a small amount of forgetting on previous tasks.

We implement AdaHAT, conduct experiments on a number of datasets and compare it with a number of task-incremental learning baselines, particularly including HAT which AdaHAT extends. Our experimental results show that AdaHAT achieves better average performance across tasks than these baselines, especially on long task sequences, demonstrating the benefits from balancing the stability-plasticity trade-off of a network when learning such sequences of tasks. This paper makes the following contributions: 
\begin{enumerate}
    \item We propose AdaHAT mechanism, which allows adaptive updates to previously static parameters in a neural network, alleviating the problem of insufficient network capacity when learning long task sequences; 
    \item We develop a neural network architecture based on the architecture developed in HAT~\cite{serra2018overcoming}, integrating our proposed task-based attention mechanism for adaptive parameter updating in the network while retaining the spirit of hard attention to the task proposed in HAT;
    \item Our experimental results show that, while slightly causing forgetting, AdaHAT improves the average performance across tasks  after the network has reached its capacity limit, which effectively balances the stability-plasticity trade-off in continual learning.
\end{enumerate}

\section{Related Work}

Task-incremental learning is a continual learning scenario where the algorithm learns a sequence of distinct tasks in an incremental manner~\cite{van2019three}. In this scenario, it is crucial to balance the stability-plasticity trade-off to ensure the optimal model performance across all tasks~\cite{wang2024comprehensive}.

Catastrophic forgetting is a major problem in task-incremental learning, where neural networks tend to overwrite knowledge learned in previous tasks when trained on new tasks. Major efforts have been focused on developing various mechanisms to prevent the network from catastrophic forgetting, often leveraging certain forms of information about previous tasks~\cite{wang2024comprehensive}. In replay-based approaches, parts of the data from previous tasks are stored and replayed during training on new tasks to mitigate forgetting. Examples include GEM~\cite{lopez2017gradient}, DGR~\cite{shin2017continual}, and DER~\cite{buzzega2020dark}. In regularization-based approaches, forgetting is mitigated by introducing regularization terms into the loss function, usually constructed from the information about previous tasks. Examples include LwF~\cite{li2017learning}, EWC~\cite{kirkpatrick2017overcoming}, SI~\cite{zenke2017continual}, IMM~\cite{lee2017overcoming}, and VCL~\cite{nguyen2017variational}. These approaches aim to inject stability into the network in their forgetting prevention mechanisms, but still generally lean towards plasticity in the trade-off.

Architecture-based approaches adopt distinctly different strategies that overemphasize stability, shifting the trade-off towards stability instead. They allocate different parts of a neural network to different tasks, i.e., specifying task-specific parameters, which leverages the inherent separability of neural network architectures. One strategy develops incrementally parallel subnetworks to learn the sequence of tasks, exemplified by Progressive Neural Networks~\cite{rusu2016progressive}. Other strategies allocate the parameter space within a fixed (sometimes dynamically expanded when needed) network architecture, with the allocation determined either by a set of rules, such as PackNet~\cite{mallya2018packnet}, or by trainable masks over the architecture trained along with network parameters, such as Piggyback~\cite{mallya2018piggyback}, HAT~\cite{serra2018overcoming}, CPG~\cite{hung2019compacting}, and SupSup~\cite{wortsman2020supermasks}. Architecture-based approaches generally suffer from the network capacity problem because they preserve and freeze task-specific parameters for previous tasks. In this case, the network can achieve maximal stability and the best performance when there is sufficient capacity, but has to face drastic performance degradation on new tasks when capacity runs out. In other words, they sacrifice learning plasticity for potential future tasks to maintain the stability for previous tasks~\cite{wang2024comprehensive}. 

To alleviate the network capacity problem, many architecture-based approaches introduce measures to carefully control the network capacity usage, with mechanisms such as sparsity regularization~\cite{serra2018overcoming}, which can sometimes in turn limit their ability to learn current tasks~\cite{wang2024comprehensive}. Others gain additional capacity resources by breaking the assumption of the fixed parameter space: some allow dynamically expanding the network when capacity runs out as more new tasks arrive. Some approaches like Progressive Neural Networks~\cite{rusu2016progressive} even expand the network every time when a new task arrives, causing progressively linear increasing computation and memory cost.

Additionally, most architecture-based approaches use many hyperparameters to control their network capacity usage, usually without directly leveraging information about previous tasks. These hyperparameters need to be tuned manually to determine how much capacity should be allocated to new tasks.  For example, PackNet
uses a pruning ratio to allocate a fixed proportion of network parameters to new tasks~\cite{mallya2018packnet}. HAT requires manual tuning the hyperparameter $s_{\text{max}}$, where larger values provide more stability for previous tasks, and smaller values provide more plasticity for new tasks~\cite{serra2018overcoming}. However, in real-world continual learning scenarios, one does not know how many new tasks will arrive~\cite{de2021continual}, or may even encounter an infinite sequence of tasks, making proper hyperparameter selection difficult. Even if it is known beforehand and the hyperparameters are well chosen for these tasks, network capacity can still become insufficient when additional tasks arrive, and the network capacity problem reappears.

Network capacity is a critical factor affecting learning plasticity when incrementally learning long task sequences~\cite{de2021continual}. To address this problem, we need to balance the stability-plasticity trade-off as well as prevent catastrophic forgetting at the same time. In this paper, we propose a new adaptive task-based attention mechanism to balance the trade-off in architecture-based approaches, enabling the adaptive allocation of network capacity with taking into account the information about previous tasks. Our proposed mechanism, in particular, alleviates the problem of lacking plasticity when incrementally learning long task sequences.

\section{Task-Incremental Learning with Adaptive Hard Attention to the Task}

In this section, we present a new approach for task-incremental learning, our proposed adaptive attention mechanism called Adaptive Hard Attention to the Task (AdaHAT), to balance the trade-off between stability and plasticity. Our proposed mechanism extends the mechanism called Hard Attention to the Task (HAT) proposed in \cite{serra2018overcoming}. 

In the task-incremental learning scenario, a sequence of tasks $t=1, \cdots, N$, arrive at a neural network in an incremental manner, with each task associated with dataset $D^t = \{x^t, y^t\}$. The objective of task-incremental learning for the network is to learn the task sequence, preventing performance degradation on previous tasks when learning new tasks, and eventually achieve better performance across all tasks~\cite{de2021continual}. To achieve this objective, the stability-plasticity trade-off needs to be balanced properly.

We adopt the hard attention to the task mechanism proposed in HAT, in which layer-wise attention vectors (masks) $\textbf{m}^{t}_l$ with binary values are learned to pay hard attention to the units in each layer $l=1, \cdots, L-1$ for a new task $t$. The attention vectors are gated from layer-wise task embeddings $\mathbf{e}_l^t$ with real values:
\begin{equation}
\mathbf{m}_l^t=\sigma\left(s \mathbf{e}_l^t\right),
\label{eq:mask}
\end{equation}
where $s$ is a positive scaling factor and $\sigma(\cdot)$ denotes the sigmoid gate function. The attention vectors are learned instead of tuned, as the task embeddings are trained along with network parameters. These binary attention vectors determine which part of the network is allocated to the task. We use $\textsf{M}^t$ to denote all the attention vectors to the task $t$, $\Theta$ to denote the parameter space, therefore $\Theta^t$, parameters in $\Theta$ allocated to task $t$, are those masked by $\textsf{M}^t$.

When training on a new task $t$, HAT conditions gradients in the backward pass according to cumulative attention vectors $\textbf{m}^{\le t}_l$ from all previous tasks. The cumulative attention vectors are recursively computed by
\begin{equation}
\textbf{m}^{\le t}_l = \max\left(\textbf{m}^t_l, \textbf{m}^{\leq t-1}_l\right)
\label{eq:cumulative}
\end{equation}
after learning task $t$, using element-wise maximum\footnote{$\textbf{m}^{\le 0}_l$ starts with all zeros to compute $\textbf{m}^{\le 1}_l$.}. This preserves the attention values for units in the network that are important to previous tasks, and allows these preserved values to condition network training on the new tasks. To condition training on the new task $t$, HAT modifies the gradient of the parameter $\theta_{l,ij}$ connecting the $j$-th unit in layer $l-1$ to the $i$-th unit in layer $l$, with a parameter-wise adjustment rate $a_{l,ij}$:
\begin{equation}
g'_{l,ij}= a_{l,ij} \cdot g_{l,ij},\ a_{l,ij} \in \{0, 1\},
\label{gradient_modify}
\end{equation}
where $g_{l,ij}$ is the gradient of parameter $\theta_{l,ij}$, and 
\begin{equation}
a_{l,ij}  = 1-\min\left(m^{< t}_{l,i},m^{< t}_{l-1,j}\right) 
\label{hat-clip}
\end{equation}
marks the hard clipping of the gradient, computed as the reverse of the minimum of the two cumulative attention values, which results in binary values again. This means, those parameters become static without updates when the units connected at both ends are masked by the cumulative attention vectors, as their gradients are hard clipped.

As more tasks arrive, more active parameters become static, gradually taking up more network capacity, hence reducing learning plasticity for new tasks. To address this network capacity problem, a regularization term explicitly controlling mask sparsity is employed in HAT to promote higher compactness of the masks and lower network capacity usage:
\begin{equation}
\mathcal{L}'\left(f(x_t), y_t,\textsf{M}^t,\textsf{M}^{<t}\right)=\mathcal{L}(f(x_t), y_t)+cR\left(\textsf{M}^t,\textsf{M}^{< t}\right),\label{eq:loss_L}
\end{equation}
\begin{equation}
    R\left(\textsf{M}^t,\textsf{M}^{<t}\right)=\frac{\sum_{l=1}^{L-1}\sum_{i=1}^{N_l}m_{l,i}^t\left(1-m_{l,i}^{<t}\right)}{\sum_{l=1}^{L-1}\sum_{i=1}^{N_l}\left(1-m_{l,i}^{<t}\right)},\label{eq:loss}
\end{equation}
where $c > 0$ denotes the regularization coefficient and $N_l$ denotes the number of units in layer $l$. This, to a certain extent, helps alleviate the network capacity problem. However, the network capacity will eventually run out, as parameters will become permanently static once they are learned to be allocated to previous tasks. As shown in an experiment illustrated in Figure~\ref{fig:HAT-mask}, the cumulative attention vector of certain layer is rapidly occupied by value $1$ as the network is trained on new tasks, which means the network rapidly exhausts its available parameter space, leading to insufficient network capacity for the algorithm to allocate parameters to new tasks, hence significantly affecting the performance of the network on new tasks.

\setcounter{figure}{-1}

\begin{figure}[htbp]
\subfigure{
        \begin{minipage}[b]{0.36\linewidth}
        \centering
        \includegraphics[width=0.8\linewidth]{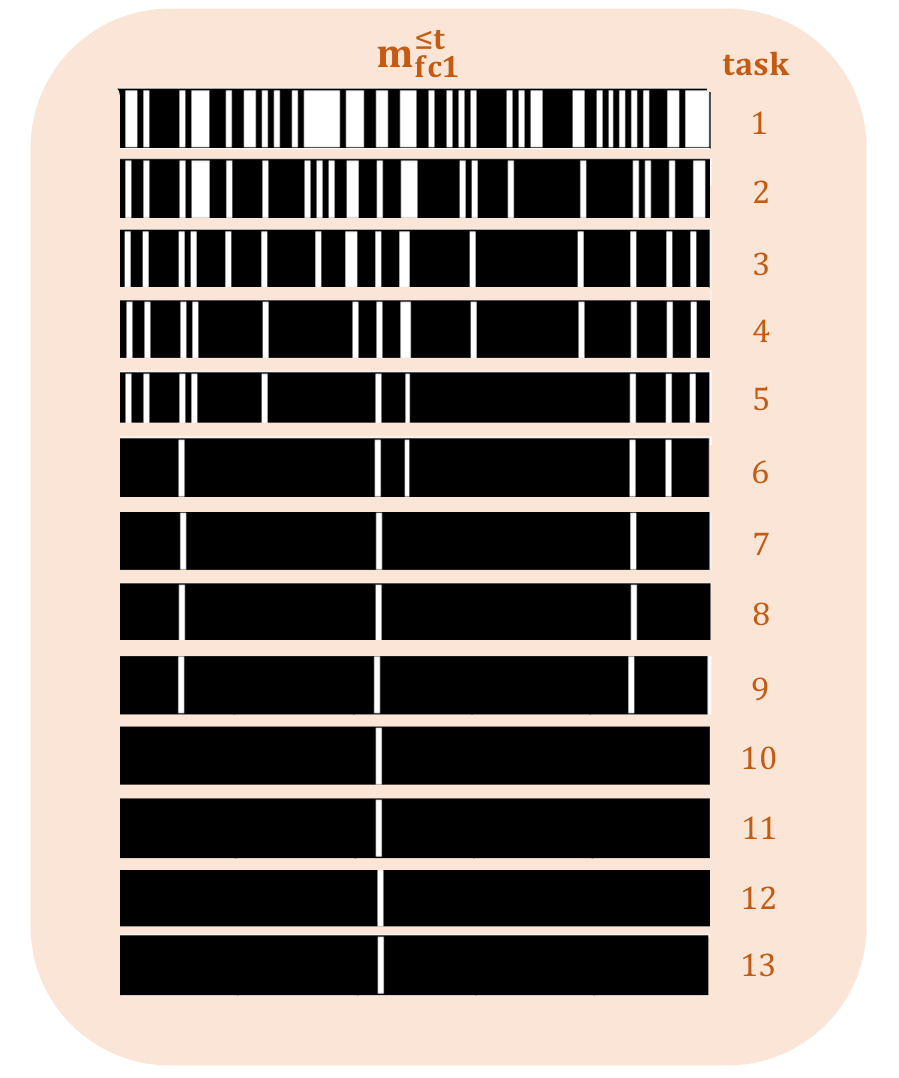} 
        \caption{Evolution of the cumulative attention vector to previous and current tasks for the first fully-connected layer (fc1) in a MLP architecture, represented by  $\textbf{m}^{\leq t}_{\text{fc}1}$, with network parameters allocated to the tasks highlighted in black.}\label{fig:HAT-mask}
        \end{minipage} 
        }
\subfigure{
        \begin{minipage}[b]{0.65\linewidth}
        \includegraphics[width=1\linewidth]{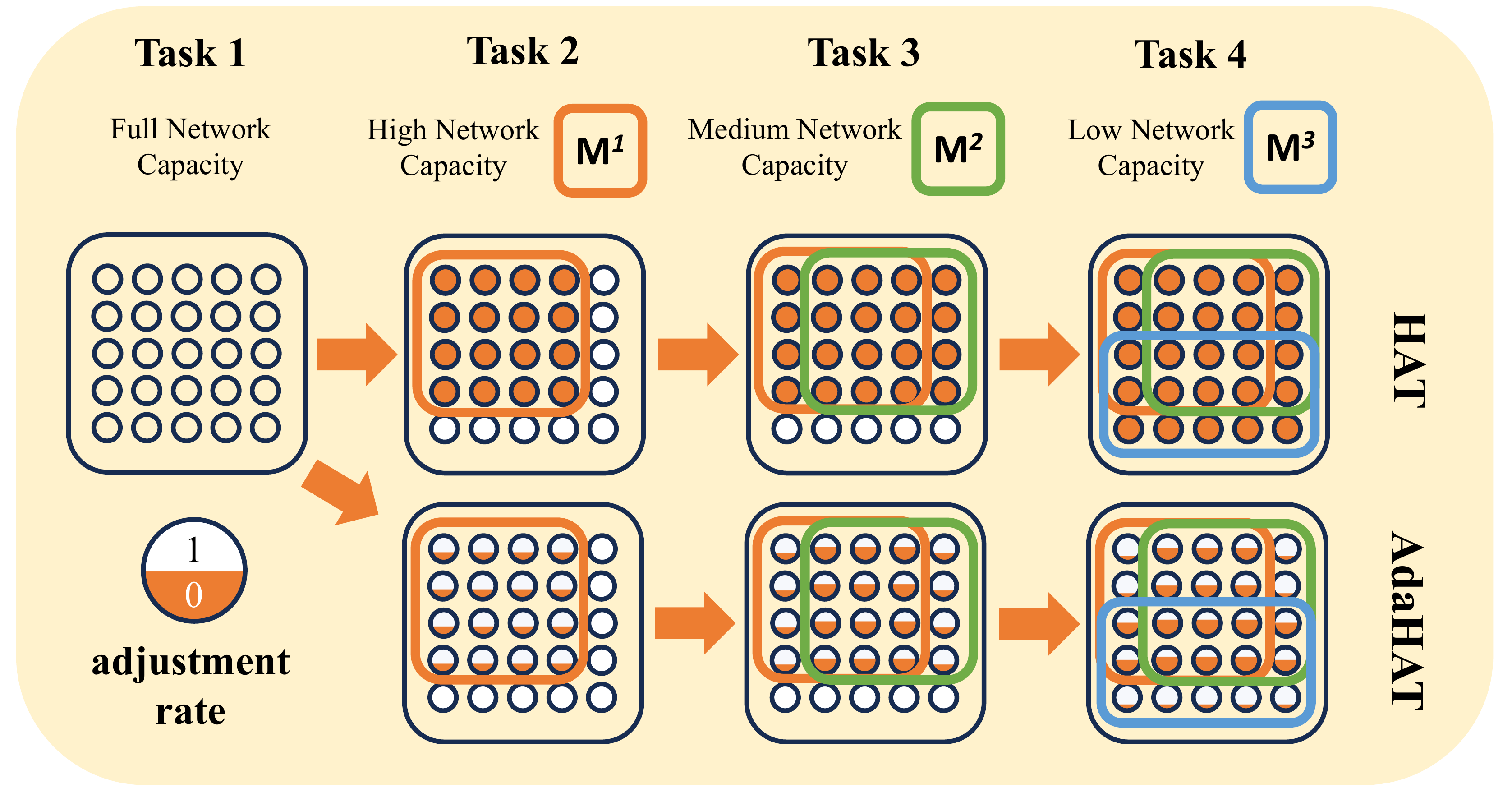} 
        \caption{Comparison of HAT and AdaHAT in the evolution of gradient adjustment rates for network parameters (represented as circles). Hard gradient clipping in HAT binarizes parameters into static (orange) and active (white) states, whereas soft gradient clipping in AdaHAT adaptively sets adjustment rates between 0 and 1 based on both parameter importance (the number of rounded rectangles enclosing a parameter) and network sparsity.}
        \label{fig:adjustment}
        \end{minipage} 
        }
\end{figure}

\subsection{The AdaHAT Algorithm: Adaptive Parameter Updating with Summative Attention to Previous Tasks}
\label{sec:AdaHAT-algorithm-section}

Following the spirit of the hard attention to the task mechanism proposed in HAT~\cite{serra2018overcoming}, in order to address the network capacity problem in HAT when learning long task sequences, we propose a new adaptive parameter updating algorithm AdaHAT based on HAT. We propose to replace the cumulative attention vectors used in HAT with our new summative attention vectors, computed as follows:
\begin{equation}
\textbf{m}^{\leq t,\text{sum}}_l = \textbf{m}^t_l +  \textbf{m}^{\leq t-1, \text{sum}}_l.
\label{eq:summative}
\end{equation}

The network capacity problem in HAT can be alleviated by allowing adaptive updates to parameters in the network that have become static for previous tasks. As shown in Eq.~\eqref{gradient_modify}, hard gradient clipping is the root cause of this problem: parameters allocated to previous tasks are strictly and permanently frozen when learning new tasks, leaving no room for future tasks to encode their knowledge in them. In AdaHAT, we therefore propose soft clipping of gradients as follows to release some capacity from these static parameters for previous tasks:
\begin{equation}
g'_{l,ij}=a^\star_{l,ij} \cdot g_{l,ij},  \  a^\star_{l,ij} \in [0, 1],
\label{eq:ada-clip}
\end{equation}
where $a^\star_{l,ij}$ denotes the adjustment rate of AdaHAT. Unlike the binary $a_{l,ij}$ in HAT, $a^\star_{l,ij}$ ranges from 0 to 1, soft clipping the gradient $g_{l,ij}$. This allows controlled updates to parameter $\theta_{l,ij}$. 

Two pieces of information about previous tasks are crucial for gradient adjustment and for determining how much network capacity should be released during training: parameter importance to previous tasks and current network capacity usage. We therefore design our adjustment rate to adapt to both. In the following, we first define two corresponding measures: \textit{parameter importance} and \textit{network sparsity}, and then incorporate them into the formulation of the adjustment rate. 

\paragraph{\textup{\textbf{Parameter Importance.}}} In HAT, cumulative attention vectors defined in Eq.~\eqref{eq:cumulative} are used to preserve parameters that are important to previous tasks. Larger cumulative attention values in these vectors indicate greater importance of the corresponding parameters; therefore, smaller updates are made to them when learning new tasks. However, this binary measure of parameter importance is often insufficiently discriminative. In AdaHAT, we replace the cumulative attention vectors with our new measure of parameter importance, called summative attention vectors, as defined in Eq.~\eqref{eq:summative}. For each parameter $\theta_{l,ij}$, we still compute the minimum of the summative attention values of the connected units at both ends, $\min\left(m^{< t, \text{sum}}_{l,i},m^{< t, \text{sum}}_{l-1,j}\right)$, as the importance of that parameter to previous tasks. This value ranges from $0$ to $t-1$, and therefore contains richer information about previous tasks. A higher summative attention value indicates that the parameter is allocated to more previous tasks, so smaller updates should be made to it when learning new tasks. For parameters with lower importance, more of their space can be released. In short, the adjustment rate should be negatively correlated with this parameter importance value.
\paragraph{\textup{\textbf{Network Sparsity.}}} The regularization term in HAT reflects the compactness of binary masks, which we use as a measure of \textit{network sparsity}. As indicated by Eq.~\eqref{eq:loss}, larger regularization values indicate that, when learning new tasks, the algorithm is paying more attention to allocating the active parameter space that has not been allocated to previous tasks. In this case, there is less need to update parameters that have already become static for previous tasks. Furthermore, this regularization value is closely related to current network capacity usage. In general, when a smaller proportion of parameters in the network is static, the regularization value tends to be larger, because more active parameters remain available for allocation. In such cases, network capacity is still sufficient, so the algorithm should avoid hastily releasing parameter space allocated to previous tasks and should instead focus on learning from currently active parameters. Therefore, the adjustment rate should be negatively correlated with this sparsity regularization value.

By incorporating network sparsity information into the adjustment rate, AdaHAT is designed to approximate HAT in early tasks as closely as possible: it applies smaller updates through a lower adjustment rate before the network reaches its capacity limit. Only when capacity becomes insufficient does it start making larger updates to parameters allocated to previous tasks. By retaining maximal stability for previous tasks before sacrificing plasticity for new tasks, AdaHAT follows the spirit of HAT and preserves its stability benefits while introducing adaptive behavior.

We now define the AdaHAT adjustment rate by incorporating both pieces of information about previous tasks:
\begin{equation}
    a^\star_{l,ij} = \frac{r_l}{\min\left(m^{< t, \text{sum}}_{l,i},m^{< t, \text{sum}}_{l-1,j}\right)+r_l},\ r_l = \frac{\alpha}{R\left(\textsf{M}^t,\textsf{M}^{<t}\right) + \epsilon},
\label{adjust_rate}
\end{equation}
where $\alpha > 0$ is a hyperparameter that controls the overall intensity of gradient adjustment. The constant $\epsilon$ is set to a small value to avoid division by zero. Note that, for active parameters, the adjustment rate is 1 because their parameter importance (summative attention) value is $0$; therefore, soft clipping also applies only to gradients of parameters allocated to previous tasks. As shown by the formulation, the adjustment rate is negatively correlated with both parameter importance and network sparsity. These two pieces of information jointly control gradient adjustment and adaptively manage network capacity usage based on previous tasks and the current learning state. We present an illustration of how the adjustment rates evolve as new tasks arrive in Figure~\ref{fig:adjustment} and summarize the AdaHAT algorithm in Algorithm~\ref{algo:adahat}.

\begin{algorithm}[htbp]
\caption{Adaptive Hard Attention to the Task (AdaHAT)}
\KwIn{task sequence $D^t, t =1,2,3, \cdots$; adjustment intensity $\alpha$; HAT hyperparameters $c$, $s_\text{max}$; learning rate $\eta$. }
\KwOut{trained network parameters $\Theta$ shared by all tasks; hard attention vectors (binary mask) $\textsf{M}^t$ for each task $t=1,2,3,\cdots$}
\BlankLine
\textbf{Initialize:} task embeddings $\textbf{e}^{t}_{l}$ from $\mathcal{N}(0,1)$; summative and cumulative attention vectors $\textbf{m}^{\leq 0, \text{sum}}_l$, $\textbf{m}^{\leq 0}_l$ both to zeros.
\BlankLine
\For{task $t = 1, 2, \cdots$}{
    \For{each training step}{

        Forward propagate input through the network under mask $\textbf{m}^{t}_{l}$ using Eq.~\eqref{eq:mask};
        
        Compute current network sparsity $R$ using Eq.~\eqref{eq:loss};
        
        Compute loss $\mathcal{L}'$ using Eq.~\eqref{eq:loss_L} and backpropagate gradients $g_{l,ij}$ of network parameters $\theta_{l,ij}$ and gradients of task embeddings $\textbf{e}^{t}_{l}$;
        
        Compute adjustment rate $a^\star_{l,ij}$ using Eq.~\eqref{adjust_rate};
        
        Soft-clip the gradients to $g'_{l,ij}$ using Eq.~\eqref{eq:ada-clip} with the adjustment rate; update $\theta_{l,ij}$ using adjusted gradients $g'_{l,ij}$ and update $\textbf{e}^{t}_{l}$ with learning rate $\eta$; 
    }
    Compute the summative and cumulative attention vectors $\textbf{m}^{\leq t, \text{sum}}_l$, $\textbf{m}^{\leq t}_l$  of task $t$ using Eq.~\eqref{eq:summative} and Eq.~\eqref{eq:cumulative}.
}
\label{algo:adahat}
\end{algorithm}

\section{Experiments}

In this section, we first introduce our experimental setup and metrics for performance, stability-plasticity trade-off and network capacity usage. We then present our experimental results of the performance on long task sequences and how AdaHAT achieves it by balancing stability-plasticity trade-off. Finally, we present ablation study of two pieces of information and hyperparameter study. Our findings demonstrate that the proposed adaptive parameter updating approach, which takes into account both the parameter importance and the current network capacity usage, is effective. It can make adaptive gradient adjustments when learning new tasks, within a very small adjustment rate.

\subsection{Experimental Setup}

\paragraph{\textbf{\textup{Task Sequences.}}}
AdaHAT aims to address the problem of insufficient network capacity when learning long sequences of tasks. Our experiments are conducted on long task sequences containing 20 tasks, which are longer than those commonly used in conventional continual learning experiment setups, where their task sequences usually contain fewer than 10 tasks. This long task sequence setting allows us to better observe how architecture-based approaches behave once the network reached its capacity limit and how performance evolves afterward, thereby exposing the network capacity problem more clearly. We also conduct experiments on much longer sequences of up to 50 tasks, which further increases the difficulty of the setting. 

\paragraph{\textbf{\textup{Data.}}}
We use the permuted version of MNIST~\cite{srivastava2013compete} and the split version of CIFAR-100~\cite{kaushik2021understanding} as our experiment datasets. We do not use permuted versions of  CIFAR-10 and CIFAR-100 because, under long task sequences, they are overly challenging for most baselines, resulting in uniformly low performance and making meaningful comparisons difficult. In addition, Split MNIST and Split CIFAR-10 are unsuitable for our setting because their 10 classes cannot be partitioned into long task sequences (e.g., 20 tasks).

\paragraph{\textbf{\textup{Baselines.}}}
To evaluate the performance of AdaHAT, we compare it with the following baselines: 
\begin{itemize}
    \item \textbf{Finetuning:} standard gradient-based training with no mechanism to prevent forgetting~\cite{goodfellow2013empirical};
    \item \textbf{Freezing:}  Stops learning after the first task;
    \item \textbf{LwF:} Learning without Forgetting~\cite{li2017learning};
    \item \textbf{EWC:} Elastic Weight Consolidation~\cite{kirkpatrick2017overcoming};
    \item \textbf{HAT:} Hard Attention to the Task~\cite{serra2018overcoming}.
\end{itemize}

We also compare AdaHAT with other gradient adjustment strategies. These strategies adjust gradients under the HAT architecture in naive or less meaningful ways (randomly or uniformly), without guidance by information about previous tasks:
\begin{itemize}
    \item \textbf{HAT-random:} HAT with an adjustment rate of random value between 0 and 1, where $a_{l,ij}$ in Eq.~\eqref{gradient_modify} is replaced by $
\begin{cases}
    \text{rand}(0,1), & \text{if } 1-\min\left(m^{< t}_{l,i},m^{< t}_{l-1,j}\right) = 0, \\
    1, & \text{otherwise.}
\end{cases}
$
\item \textbf{HAT-const-alpha:} HAT with a fixed adjustment rate of constant value $\alpha$ (equals to $\alpha$ in Eq.~\eqref{adjust_rate}), where $a_{l,ij} = 
\begin{cases}
    \alpha, & \text{if } 1-\min\left(m^{< t}_{l,i},m^{< t}_{l-1,j}\right) = 0, \\
    1, & \text{otherwise.}
\end{cases}
$
\item \textbf{HAT-const-1:} HAT with a fixed adjustment rate of constant value $1$, meaning that no gradient flow is blocked by previous task masks during backpropagation. The subnetwork for the current task is completely free to learn, regardless that it contains parameters allocated to previous tasks.
\end{itemize}

\paragraph{\textbf{\textup{Evaluation Metrics.}}} We evaluate AdaHAT and the baselines using 4 metrics: performance, stability, plasticity and network capacity usage. We compute Average Accuracy (AA)~\cite{wang2024comprehensive} and Forgetting Ratio (FR)~\cite{serra2018overcoming} on each dataset over all tasks, which are the main performance metrics in task-incremental learning. In addition, to characterize the stability-plasticity trade-off and reveal whether each approach tends to favor stability, plasticity, or a more balanced behavior, we report Backward Transfer (BWT)~\cite{wang2024comprehensive} and Forward Transfer (FWT)~\cite{wang2024comprehensive} to measure stability and plasticity, respectively.

Let $a_{t,N}$ denote the accuracy of the test model on dataset $D^t$ after learning task $N$, $a^J_{t,N}$ denote the accuracy on dataset $D^t$ of a randomly-initialized reference model jointly trained on $\cup_{\tau=1}^N D^\tau$, $a^I_t$ and  $a^R_t$ denote the accuracy of a randomly-initialized reference model independently trained on $D^t$ and a random stratified model, respectively. The metrics are defined as follows~\cite{wang2024comprehensive}:
\begin{equation}
    \mathrm{AA}_N=\frac{1}{N} \sum_{t=1}^N a_{t, N},
\end{equation}
\begin{equation}
    \mathrm{FR}_N=\frac{1}{N} \sum_{t=1}^N \frac{a_{t, N}-a^R_{t}}{a^J_{t, N}-a^R_{t}} - 1,
\end{equation}
\begin{equation}
    \mathrm{BWT}_N=\frac{1}{N-1} \sum_{t=1}^{N-1}\left(a_{t, N}-a_{t, t}\right),
\end{equation}
\begin{equation}
    \mathrm{FWT}_N=\frac{1}{N-1} \sum_{t=2}^N\left(a_{t, t}-a^I_t\right).
\end{equation}
Each experiment is repeated 5 times, and we report the mean and standard deviation for each metric as percentages.

We propose to measure network capacity usage by the relative degree to which parameters are allowed to be updated, as reflected by the adjustment rates\footnote{AdaHAT's adjustment rates are denoted as $a^\star_{l,ij}$.}:
\begin{equation}
    \text{NC} = \frac{1}{\sum\limits_{l,i,j} 1}\sum_{l,i,j} a_{l,ij}.
\end{equation}
The values of NC ranges from $0$ to $1$, where 1 indicates that all parameters can be updated freely with no adjustment to their gradients, and 0 indicates that all gradients are set to zeros and no parameter is allowed to be updated.

\paragraph{\textbf{\textup{Networks.}}}
For experiments on Permuted MNIST, we use a fully-connected (MLP) network architecture including 3 hidden layers with dimensions 256, 100, and 64 as the feature extractor. For the experiments on Split CIFAR-100, we use ResNet-18~\cite{he2016deep} as the feature extractor. Task embeddings are integrated into each layer to generate task-based attention vectors~\cite{serra2018overcoming}. Rectified linear unit is used as the activation function. As this is a task-incremental learning scenario, each task has its own output head which is a linear output layer, and the number of heads increases as new tasks arrive. All layers are randomly initialized, except for task embeddings $\mathbf{e}^t_l$, which are initialized from a standard normal distribution $\mathcal{N}(0, 1)$.

\paragraph{\textbf{\textup{Hyperparameters and Training Details.}}} We set the adjustment intensity $\alpha$ in Eq.~\eqref{adjust_rate} to $10^{-6}$, as discussed in Section~\ref{sec:hyperparameters}. $\epsilon$ in Eq.~\eqref{adjust_rate} is set to $0.1$.   Since the experiments are intended to study the network after it has reached the capacity limit, the hyperparameter $s_\text{max}$ in HAT is set to a large value of $400$; $c$ is set to $0.1$. 

All models are trained using the Adam optimizer~\cite{kingma2014adam} with a learning rate of $0.001$. We apply gradient clipping with a threshold of 0.001 and weight decay with a coefficient of 0.00035 to all approaches except those based on the HAT architecture, since gradient clipping and weight decay are incompatible with HAT. We train for 2 and 20 epochs for each task in Permuted MNIST and Split CIFAR-100. The batch sizes are set to 128 and 64, respectively. Training for Permuted MNIST is conducted on local CPUs, whereas training for Split CIFAR-100 is conducted on a server cluster equipped with NVIDIA RTX A4000 GPUs (16 GB VRAM). Our code is implemented as an open-source Python package for continual learning research\footnote{\url{pengxiang-wang.com/projects/continual-learning-arena}}.

\subsection{Main Results and Analysis}

\begin{table}[t]
% \footnotesize
\small
\centering
\caption{Results on performance and stability-plasticity trade-off metrics (mean $\pm$ std) of different approaches on the two datasets (20 tasks). }
\resizebox{\textwidth}{!}{
\begin{tabular}{llcccc} 
\toprule[1.5pt]
Dataset      & Approach              &AA($\uparrow$)    & FR ($\uparrow$)                & BWT       &FWT  \\ \midrule[1pt]
\multirow{8}{*}{\parbox{1.5cm}{Permuted\\MNIST}}  
            &Finetuning       & $32.62\pm1.60$  & $-73.78 \pm 1.84$ &   $-68.10  \pm 1.68$ &    $0.10 \pm 0.04$              \\ 
            &Freezing         & $14.73 \pm 0.48$ & $-94.04 \pm 0.70$ & $0.00 \pm 0.00$ & $-87.13 \pm 0.53$ \\
            & LwF                   & $26.95\pm 1.80$  & $-80.35  \pm  2.08$   & $-72.59 \pm 1.91$   & $ -0.04 \pm 0.04 $  \\ 
            & EWC                   & $52.25 \pm2.46$  & $-51.38  \pm 2.83$  & $-42.04   \pm 2.67$    & $-8.86   \pm 0.09 $   \\
            & HAT                   & $67.64  \pm 1.27$  & $-33.70 \pm  1.46$  & $-0.11 \pm 0.18$    & $-30.54 \pm 0.27$     \\ 
            & HAT-random           & $66.43 \pm1.21$ & $-35.10   \pm 1.39$  &  $-0.27 \pm 0.49$   &  $-1.47 \pm  0.05$  \\ 
    & HAT-const-alpha      &  $68.08  \pm 1.18$  &  $-33.20\pm 1.36$ &   $-1*e^{-3} \pm 0.00$    & $-1.39 \pm 0.04$ \\ 
    & HAT-const-1      & $48.83 \pm 4.35 $ & $-55.14 \pm 5.02$  & $-49.68 \pm  4.40 $&$-3.14 \pm 0.07$  \\ 
    & \textbf{AdaHAT}               &  $\mathbf{79.90 \pm 2.40}$ & $\mathbf{-19.43 \pm 2.76}$       &   $-14.68  \pm 2.48  $    &$ -2.49 \pm0.06$ \\ \midrule[1pt]
\multirow{8}{*}{\parbox{1.5cm}{Split\\CIFAR-100}}   
&Finetuning       & $24.34\pm0.73$ & $-91.66\pm1.32 $& $-54.00\pm1.00$  & $2.61 \pm 0.32$ \\ 
&Freezing & $23.22 \pm 0.96$ & $-94.73 \pm 2.13$ & $0.00 \pm 0.00$ & $-57.53 \pm 1.20$ \\
& LwF                   & $34.56 \pm 0.94$ & $ -70.91\pm 2.05$ & $-48.03\pm 1.01$ &  $4.23 \pm 0.42$  \\ 
& EWC                   & $30.23\pm 1.61 $& $ -79.84 \pm 3.13$& $-54.05 \pm 1.28$ & $-9.80 \pm 1.92$ \\ 
& HAT                   & $32.44 \pm 1.58$& $-74.71\pm 3.37 $& $-45.59\pm 1.49$ &  $-18.75 \pm 0.94$ \\ 
& HAT-random           & $31.41 \pm 1.29$&$ -76.98\pm 2.45$&  $-48.80\pm 1.33$&  $-7.18 \pm 1.14$ \\ 
& HAT-const-alpha      & $32.16 \pm 2.48$& $-75.04 \pm 5.16$& $-44.49\pm 2.57$ & $-5.45 \pm 0.82$ \\ 
& HAT-const-1         &  $32.40 \pm 1.40$& $-75.58\pm 3.08$ &  $-48.80\pm 1.72$& $-7.14 \pm 1.12$ \\ 
& \textbf{AdaHAT} & $\mathbf{38.74 \pm 2.24}$&$\mathbf{-62.37 \pm 4.64}$&$-42.11\pm2.02$ &  $-5.49 \pm 0.57$ \\ 
\bottomrule[1.5pt]
\end{tabular}
}
\label{tb:result}
\end{table}

We present the main experimental results on sequences of 20 tasks in Table~\ref{tb:result}. Our proposed approach, AdaHAT, achieves the best overall continual learning performance in terms of AA and FR across all datasets, outperforming HAT and other baseline approaches. This demonstrates the superiority of AdaHAT for task-incremental learning on long task sequences.

We observe that many baseline approaches exhibit extreme BWT and FWT values. Finetuning, LwF, and EWC tend to favor plasticity, showing relatively low BWT and high FWT, whereas Freezing and HAT tend to favor stability, showing relatively high BWT and low FWT, both of which demonstrate an imbalanced stability-plasticity trade-off. AdaHAT, by contrast, maintains relatively balanced BWT and FWT values, with neither being excessively high nor low, indicating a more balanced stability-plasticity trade-off. Together with the AA and FR results, it suggests that stronger overall performance is often associated with an effective balance between BWT and FWT. Therefore, balancing the stability-plasticity trade-off is important for achieving the objective of continual learning, and AdaHAT is better at this.

For the HAT variants with other gradient adjustment strategies, we observe that almost none of them outperform HAT itself, let alone AdaHAT. These approaches apply gradient adjustment in a manner similar to AdaHAT, in that way they allow parameters allocated to previous tasks to be updated, but without proper guidance by information about previous tasks. This indicates that the gradient adjustment must be properly guided in order to play a beneficial role; without being guided by meaningful information about previous tasks, it can even lead to performance degradation. One notable example is HAT-const-1, which allows full parameter updates and shifts the trade-off heavily towards plasticity as reflected by its low BWT, likely contributing to its poor AA and FR performance. Overall, these findings suggest that the two pieces of information about previous tasks incorporated in AdaHAT are crucial. We discuss their individual contributions in Section~\ref{sec:adahat-ablation}.

\begin{figure}[htbp]
\resizebox{\textwidth}{!}{
\subfigure{
        \begin{minipage}[b]{0.5\linewidth}
        \includegraphics[width=1\linewidth]{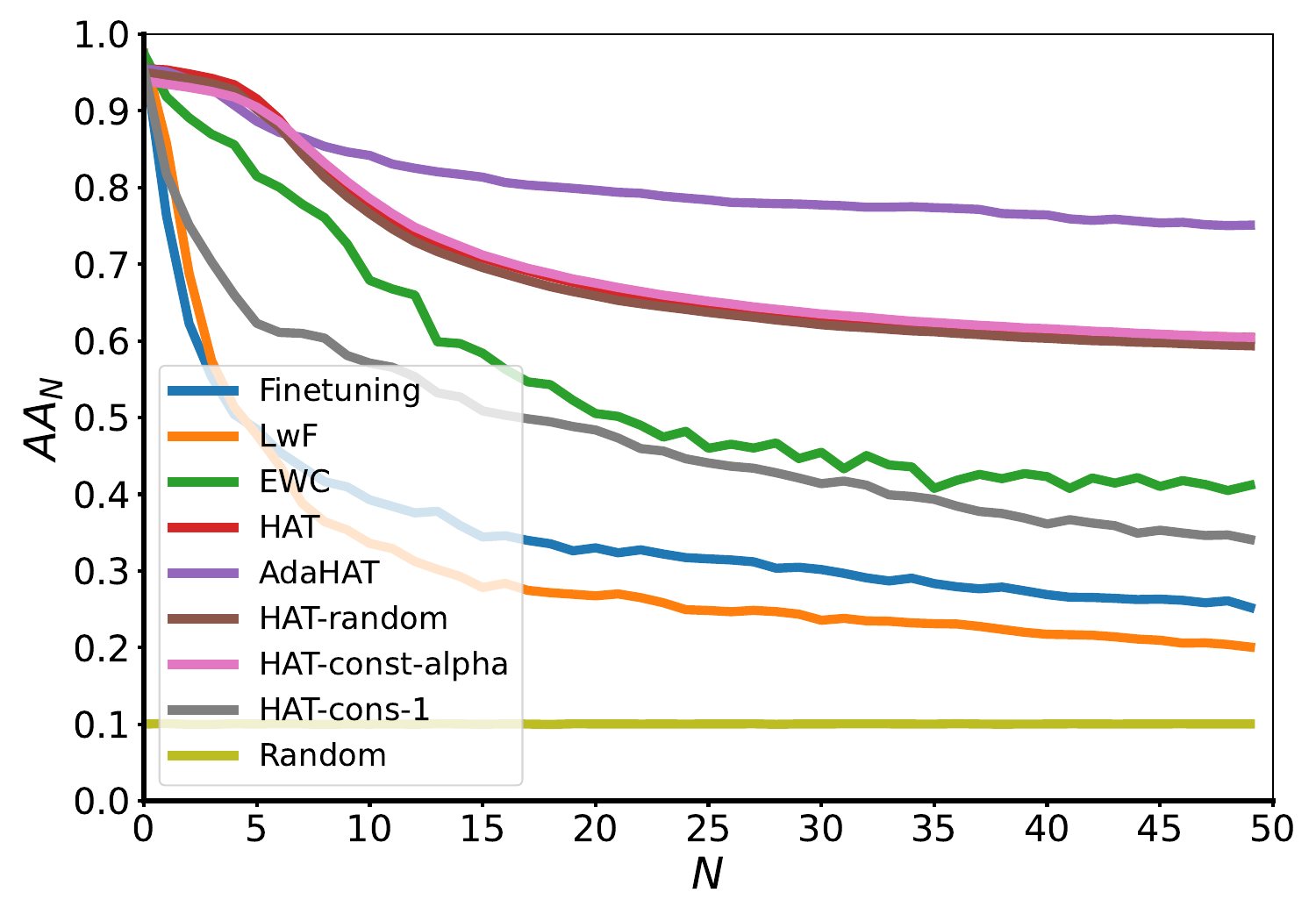} 
        \caption{Evolution of AA (mean) over the task sequence for different approaches on longer sequences of tasks (50 tasks, Permuted MNIST).}\label{fig:50tasks}
        \end{minipage} 
        }
\subfigure{
        \begin{minipage}[b]{0.5\linewidth}
        \includegraphics[width=1\linewidth]{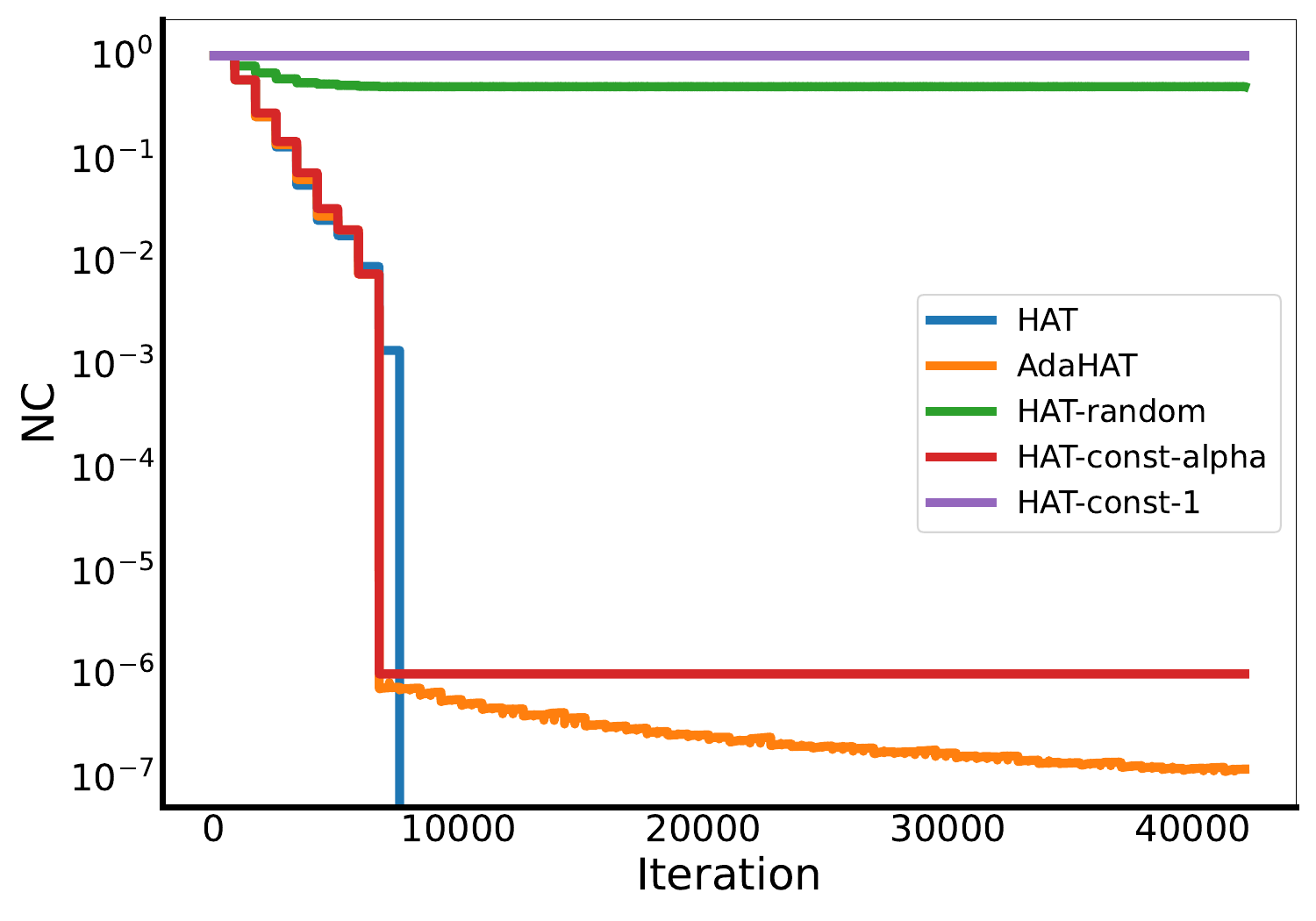} 
        \caption{Evolution of NC over training iterations for different HAT variants on longer sequences of tasks (50 tasks, Permuted MNIST).}\label{fig:capacity}
        \end{minipage} 
        }
}
\end{figure}

\paragraph{\textbf{\textup{Longer Task Sequences.}}}
To evaluate AdaHAT in a more challenging setting, where the network capacity problem becomes more severe, we conduct experiments on a much longer sequence of 50 tasks on Permuted MNIST. The performance evolution of AdaHAT and the baseline approaches over the task sequence is shown in Figure~\ref{fig:50tasks}. HAT demonstrates slightly better performance before 8 tasks, as it mitigates forgetting by strictly freezing the parameters allocated to previous tasks. However, the performance curve of HAT reaches a turning point at around the 8th task and then drops drastically. This corresponds exactly to Figure~\ref{fig:HAT-mask}, where HAT has nearly exhausted its active parameter space and the network capacity problem begins to take effect. By contrast, AdaHAT maintains a clear advantage over HAT and the other baselines after this turning point. This is because AdaHAT releases part of the allocated network capacity to make room for future task learning, thereby alleviating the network capacity problem and improving the average performance across tasks. These results further demonstrate the superiority of AdaHAT for continual learning on long task sequences.

Furthermore, we observe that although AdaHAT does not outperform HAT before 8 tasks, it behaves very closely to HAT. This validates the role of the network sparsity information: before network capacity is exhausted, it encourages AdaHAT to approximate HAT by restricting updates to parameters allocated to previous tasks to a small scale through smaller adjustment rates, thereby mitigating early forgetting and improving early-stage performance. This finding supports the discussion in Section~\ref{sec:AdaHAT-algorithm-section}.

\paragraph{\textbf{\textup{Network Capacity Usage.}}} We plot the evolution of Network Capacity (NC) usage over training iterations under 50 tasks, Permuted MNIST (seed 1) in Figure~\ref{fig:capacity}. Throughout training on this long task sequence, the network capacity of HAT is rapidly exhausted at an early stage. For the HAT variants that allow parameters allocated to previous tasks to be updated (HAT-random, HAT-const-alpha, and HAT-const-1), the capacity eventually stabilizes at values of $0.5$ (the mean value of $\text{rand}(0,1)$), $10^{-6}$ (the value to which we set $\alpha$), and $1$, respectively. AdaHAT behaves very similarly to HAT in the early stage. However, whereas HAT eventually exhausts its network capacity, AdaHAT adaptively manages capacity usage over time through its adaptive adjustment rates, causing NC to approach zero without ever reaching it, always preserving some network capacity for learning future tasks. This is the key mechanism by which AdaHAT balances the stability-plasticity trade-off when maintaining stability begins to compromise plasticity, thereby alleviating the network capacity problem in HAT.

\begin{figure}[htbp]
\resizebox{\textwidth}{!}{

    \subfigure{
        \begin{minipage}[b]{0.5\linewidth}
        \includegraphics[width=1\linewidth]{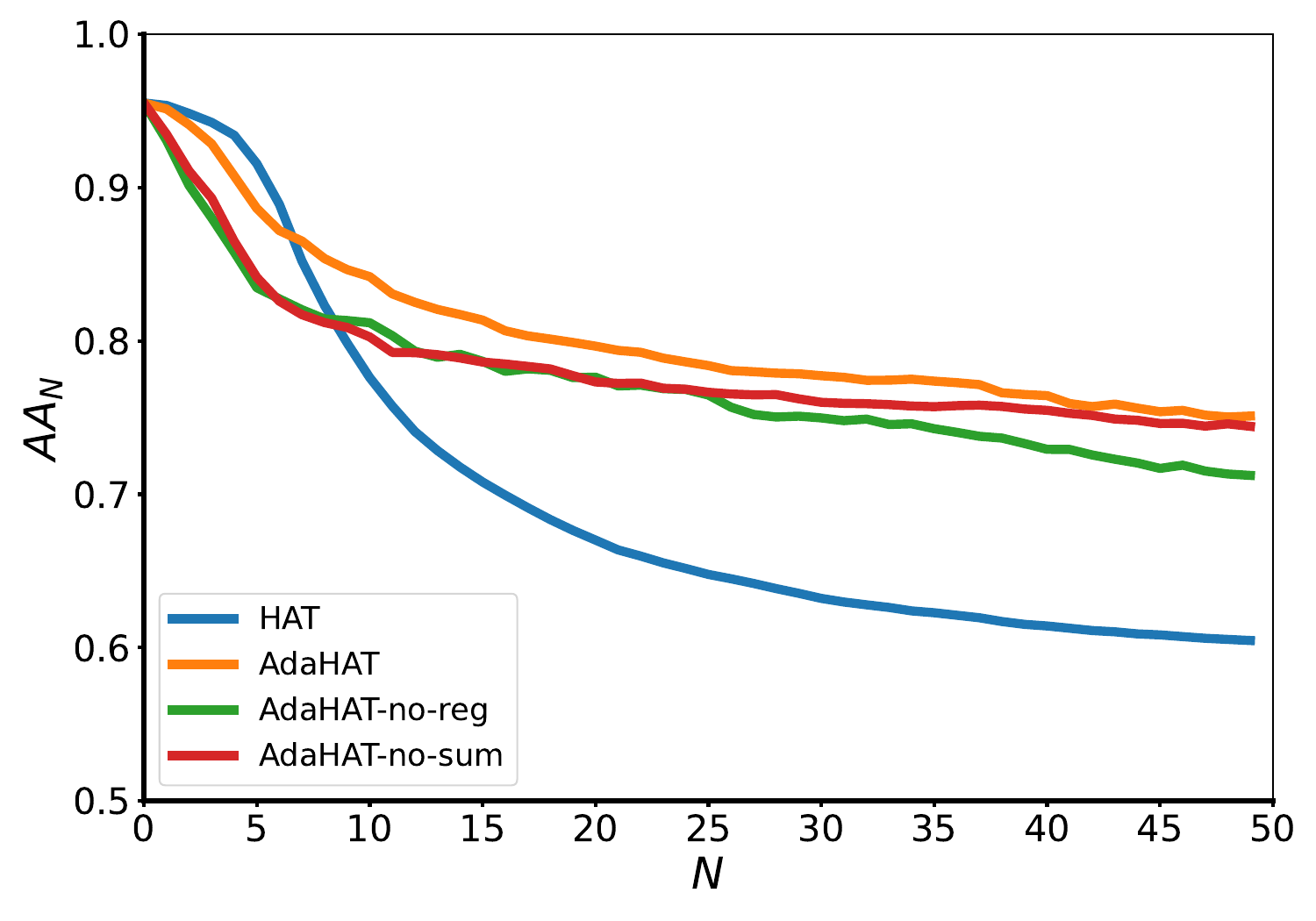} 
        \caption{Evolution of AA over the task sequence for AdaHAT and its ablated approaches on 50 tasks, Permuted MNIST.} 
        \label{fig:ablation}
        \end{minipage}
        }
    \subfigure{
        \begin{minipage}[b]{0.5\linewidth}
        \includegraphics[width=1\linewidth]{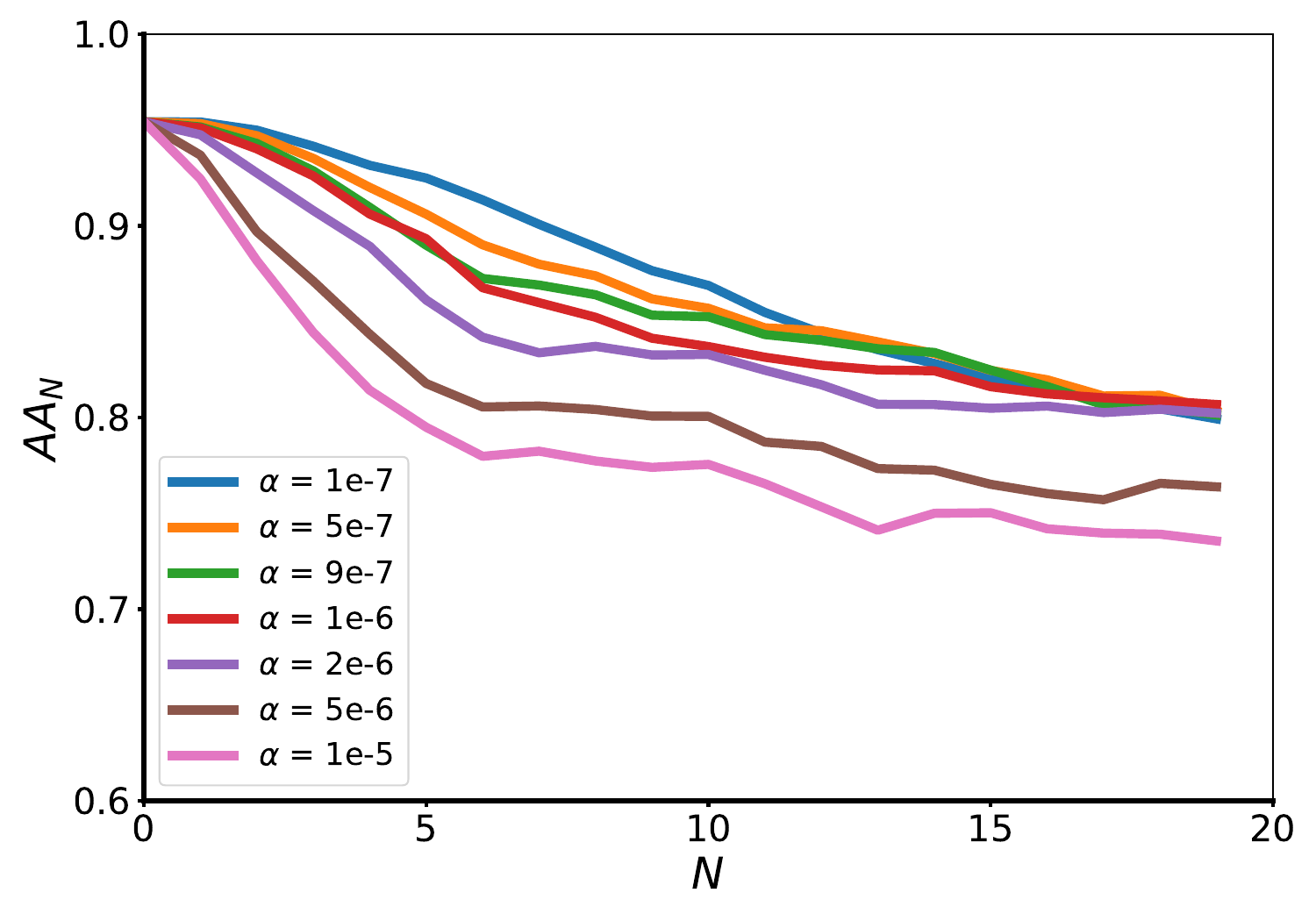} 
        \caption{Evolution of AA over the task sequence for AdaHAT with different choices of the hyperparameter $\alpha$.}
        \label{fig:hparams}
        \end{minipage} 
        }
}
\end{figure}

\subsection{Ablation Study}
\label{sec:adahat-ablation}

To provide insights into the individual effects of the two pieces of information incorporated into the adjustment rate of AdaHAT, we design and compare the following ablated variants to analyze how each of them helps improve HAT by fixing the other information as a constant value:
\begin{itemize}
    \item \textbf{AdaHAT-no-sum:} To study the effect of parameter importance, we fix all the summative attention $\min\left(m^{< t, \text{sum}}_{l,i},m^{< t, \text{sum}}_{l-1,j}\right)$ at a constant value $t$, so the adjustment rate solely depends on the network sparsity. Note that we use $t$ instead of $1$ because we want to keep the same increasing magnitude as the summative attention. In other words, AdaHAT-no-sum always treats all previous tasks with the same high level of importance.
    \item \textbf{AdaHAT-no-reg:} To study the effect of network sparsity, we fix the regularization term $R\left(\textsf{M}^t,\textsf{M}^{<t}\right)$ at a constant value 0, so the adjustment rate solely depends on the summative attention. In other words,  AdaHAT-no-reg always treats the network as if it is under insufficient sparsity. 
\end{itemize}

Figure~\ref{fig:ablation} shows the results of the ablation study under the same setting as the previous experiments on longer task sequences. Both ablated approaches fail to outperform AdaHAT, suggesting that the guidance from both pieces of information about previous tasks for the adjustment rate is crucial. Moreover, they both outperform HAT in long task sequences, suggesting that the adaptive gradient adjustment mechanism is the key to the improved overall performance of AdaHAT. 

We observe that AdaHAT-no-reg consistently underperforms AdaHAT. AdaHAT-no-reg uses more aggressive adjustment rates from the beginning, as if network capacity were insufficient, even though it is in fact still sufficient. It thus prematurely sacrifices the stability inherited from HAT before plasticity is actually reduced, leading to significant performance drops on early tasks and making it difficult to catch up with AdaHAT on future tasks. This conversely reflects the effect discussed in Section~\ref{sec:AdaHAT-algorithm-section} that the network sparsity information helps mitigate the early forgetting of AdaHAT. Similarly, we observe that AdaHAT-no-sum underperforms AdaHAT in much the same way as AdaHAT-no-reg, as it loses another piece of crucial information, leading to less effective and targeted gradient adjustments. An interesting observation here is that AdaHAT-no-sum's performance declines increasingly slowly and gradually approaches that of AdaHAT as more tasks arrive. The explanation is that, over a long task sequence, the network keeps allocating its parameters to new tasks, so that all parameters tend to have a similar likelihood of having been allocated to a comparable number of previous tasks, thereby making it increasingly difficult to distinguish parameter importance. When the parameter importance scores become less informative and discriminative, the mechanism of AdaHAT effectively reduces to that of AdaHAT-no-sum, so the two behave similarly.

\subsection{Hyperparameter Study}
\label{sec:hyperparameters}

AdaHAT introduces only one hyperparameter, $\alpha$, which acts as an additional regulation for the stability-plasticity trade-off by controlling the overall intensity of gradient adjustment.  We evaluate $\alpha$ over a range of values $10^{-7}, 2\times10^{-7}, \ldots, 9\times10^{-7}, 10^{-6}, 2\times10^{-6}, \ldots, 10^{-5}$ to determine the optimal configuration and analyze its effect.

Figure~\ref{fig:hparams} shows the results of the hyperparameter study under 20 tasks, Permuted MNIST. We observe that AdaHAT performs the best when $\alpha$ is set to $10^{-6}$ in this setting. Increasing $\alpha$ to $10^{-5}$ leads to significant performance drops as larger gradient adjustments cause more forgetting. Conversely, smaller value of $\alpha$ is not optimal either. For example, while $\alpha = 10^{-7}$ performs well in the early tasks, it drops instead after around 15 tasks. Overall, $\alpha = 10^{-6}$ achieves the optimal balance between stability and plasticity. 

Note that $10^{-6}$ is a very small value, indicating that the parameters allocated to previous tasks are restricted to extremely small updates. Since this leaves very limited room for gradient adjustment, designing a proper and well-guided adjustment rate that makes effective use of it becomes both more important and more difficult.

\section{Conclusion}

Catastrophic forgetting is one of the fundamental challenges faced by deep neural networks, which has attracted a lot of research in continual learning. Several existing architecture-based approaches that use hard attention mechanism to prevent the network from forgetting what it has learned in previous tasks tend to tilt the stability-plasticity trade-off towards stability, and suffer from the insufficient network capacity problem in long sequences of tasks. Consequently, these approaches perform well when network capacity is sufficient, yet degrade drastically once it is exhausted when learning long task sequences. In this paper, we propose a novel task-based attention mechanism, Adaptive Hard Attention to the Task (AdaHAT), which is built on HAT and replaces its hard gradient clipping with a soft, adaptive gradient adjustment that allows small and controlled updates to the parameters allocated to previous tasks. AdaHAT can preserve the stability benefits from HAT but also rebalance the stability-plasticity trade-off and alleviate the network capacity problem. Experimental results showed that AdaHAT outperforms HAT and the other baselines especially on long task sequences. The results on stability, plasticity, and network capacity usage indicate that, the adaptive behavior by which AdaHAT manages the network capacity over time, balancing the stability-plasticity trade-off, is closely related to its better performance.

Our proposed adaptive parameter updating approach also showed that in the architecture-based approaches, those static parameters allocated to previous tasks can be safely updated at small magnitudes when a long task sequence reaches the capacity limit, provided the updates are well-guided and adaptive. Concretely, both pieces of information about previous tasks we have incorporated into the adjustment rate play a crucial role in guiding these parameter updates, shown in ablation results. Going further, we believe finer-grained task information can be explored and exploited to this end. We leave this to future work.

\bibliographystyle{splncs04}
\bibliography{mybibliography}
%% Note that this preceding line implies that you store your BibTeX references in a file called 'mybibliography.bib'. If you instead store your references in a file with a different name, for instance 'references.bib', the preceding line should read '\bibliography{references}'. Whatever you do, DO NOT put the file name extension .bib inside the \bibliography command; this will trip up LaTeX compilers. 

\end{document}